\documentclass{article}
\usepackage{graphicx}

\PassOptionsToPackage{numbers, compress}{natbib}
\usepackage[dblblindworkshop, final]{neurips_2025}
\workshoptitle{SPACE in Vision, Language, and Embodied AI}

\usepackage[utf8]{inputenc} 
\usepackage[T1]{fontenc}    
\usepackage{hyperref}       
\usepackage{url}            
\usepackage{booktabs}       
\usepackage{amsfonts}       
\usepackage{nicefrac}       
\usepackage{microtype}      
\usepackage{xcolor}         
\usepackage{multirow}         
\usepackage{graphicx}
\usepackage{caption}
\usepackage{float}
\usepackage{enumitem}
\usepackage{amsmath}

\title{LayoutAgent: A Vision-Language Agent Guided Compositional Diffusion for Spatial Layout Planning}

\author{%
Zezhong Fan\thanks{Equal contribution.} \quad
  Xiaohan Li\footnotemark[1] \quad
  Luyi Ma \quad
  Kai Zhao\quad
  Liang Peng\quad \\
  \textbf{Topojoy Biswas}\quad 
  \textbf{Evren Korpeoglu}\quad
  \textbf{Kaushiki Nag}\quad 
  \textbf{Kannan Achan} \\
  Personalization Team, Walmart Global Tech\\
  Sunnyvale, California, USA \\
  \texttt{\{zezhong.fan, xiaohan.li, luyi.ma, kai.zhao, liang.peng,} \\ \texttt{topojoy.biswas, ekorpeoglu, kaushiki.nag, kannan.achan\}@walmart.com} \\
}

\begin{document}

\maketitle

\begin{abstract}

Designing realistic multi-object scenes requires not only generating images, but also planning spatial layouts that respect semantic relations and physical plausibility. On one hand, while recent advances in diffusion models have enabled high-quality image generation, they lack explicit spatial reasoning, leading to unrealistic object layouts. On the other hand, traditional spatial planning methods in robotics emphasize geometric and relational consistency, but they struggle to capture semantic richness in visual scenes. To bridge this gap, in this paper, we propose LayoutAgent, an agentic framework that unifies vision-language reasoning with compositional diffusion for layout generation. Given multiple input images with target objects in them, our method first employs visual-language model to preprocess the inputs through segmentation, object size estimation, scene graph construction, and prompt rewriting. 
Then we leverage compositional diffusion—a method traditionally used in robotics—to synthesize bounding boxes that respect object relations encoded in the scene graph for spatial layouts. 
In the end, a foreground-conditioned image generator composes the complete scene by rendering the objects into the planned layout guided by designed prompts. Experiments demonstrate that LayoutAgent outperforms other state-of-the-art layout generation models in layout coherence, spatial realism and aesthetic alignment.



\end{abstract}

\section{Introduction}

Generating realistic multi-object scenes requires not only producing high-quality images but also planning spatial layouts that adhere to semantic relations and physical plausibility~\cite{ritchie2019fast, wang2019planit}. While recent text-to-image models excel at photorealistic image generation~\cite{yu2022scaling, rombach2022high, betker2023improving, fan2024prompt}, they often lack explicit spatial reasoning, leading to implausible arrangements such as overlapping or mis-scaled objects. In contrast, traditional spatial planning methods in robotics emphasize geometric and relational consistency~\cite{xu2025set}, but they struggle to capture the semantic richness and visual fidelity needed for complex scene synthesis.

\begin{figure*}[t]
    \centering
    \includegraphics[width=1.0\textwidth]{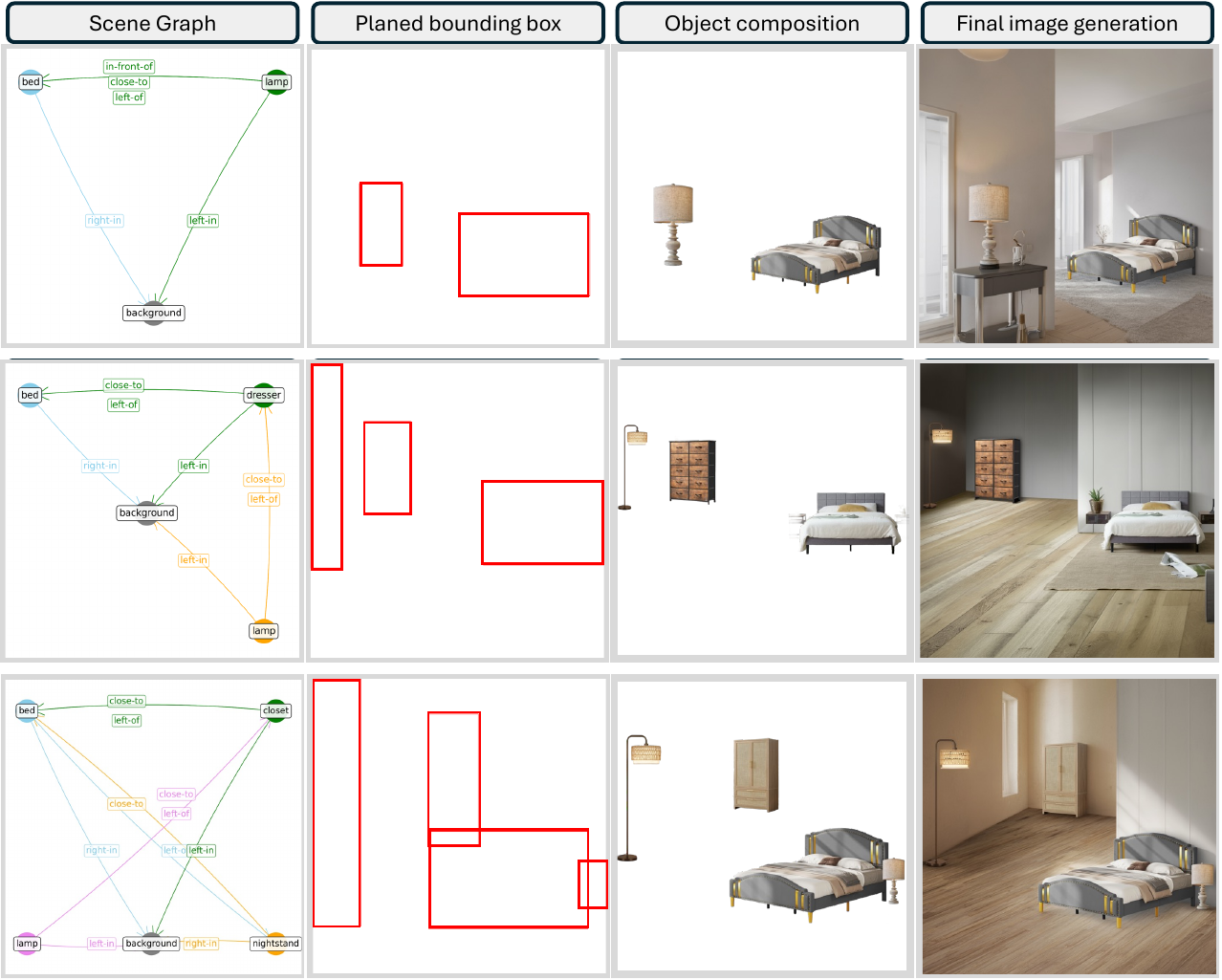}
    \caption{Qualitative results of the generated bedroom scenes with {2, 3 and 4} objects.}
    \label{fig:good_cases}
\end{figure*}

This gap motivates the need for a framework that unifies semantic reasoning with structured spatial planning. We propose LayoutAgent, an agentic framework integrates vision-language reasoning with compositional diffusion for layout generation (see Figure~\ref{fig:good_cases}). Given multiple input object images, a Visual-Language Model (VLM) agent preprocesses the inputs by performing segmentation, object size estimation, scene graph construction, and prompt rewriting. The enriched prompt and scene graph provide semantic and relational structure for subsequent layout generation. To generate layouts, we adapt compositional diffusion~\cite{liu2022compositional,xu2025set,yang2023compositional} from robotics field to generate bounding boxes that respect object relations and estimated sizes. Each pairwise relationship is modeled by an individual diffusion model, enabling fine-grained spatial reasoning. In the end, a foreground-conditioned image generator renders the complete scene, combining realistic object appearances with planned layouts. 

Our contribution can be summarized as the following points:
\begin{itemize}
    \item We propose LayoutAgent, an agentic framework integrates vision-language reasoning with compositional diffusion for layout generation.
    \item Experimental results on structured scene generation benchmarks demonstrate that LayoutAgent significantly improves \emph{Layout Coherence, Spatial Realism, and Aesthetic Alignment} over strong baselines.
    \item The ablation studies confirm the complementary benefits of compositional diffusion and scene graph planning. In case studies, we illustrate its systematic, multi-stage planning for scene generation.
\end{itemize}

\section{Methodology}

\begin{figure*}[ht]
    \centering
        \includegraphics[width=\textwidth]{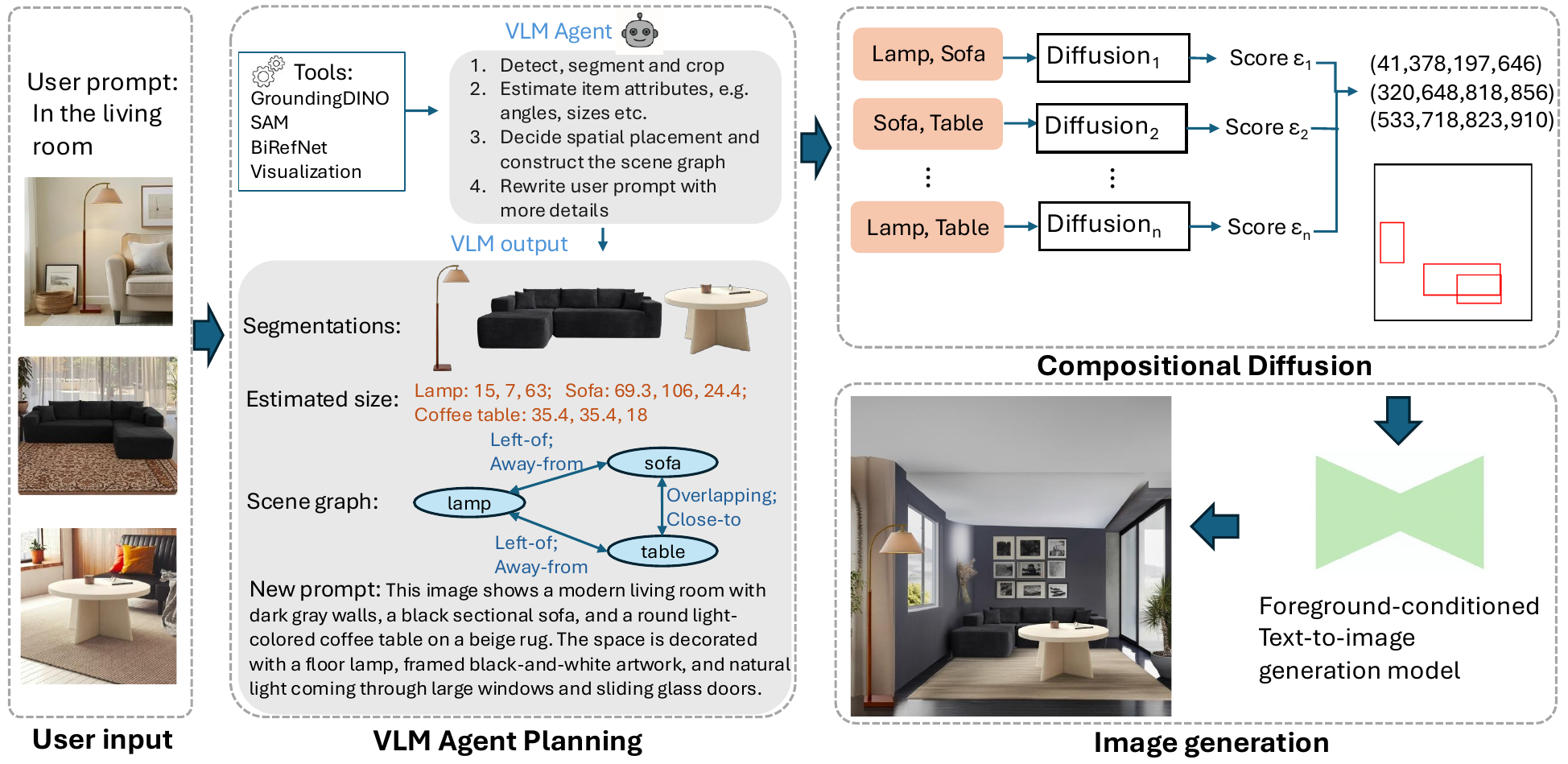}
        \caption{Our Overall Framework.}
        \label{fig:framework}
\end{figure*}
In this section, we present LayoutAgent, a unified framework that combines vision-language reasoning with compositional diffusion to generate spatial layouts that are both visually realistic and physically plausible across diverse scenes. As shown in Figure~\ref{fig:framework}, our approach integrates high-level reasoning and spatial planning with generative modeling through three core components: (1) a Vision-Language Agent that interprets input objects and produces high-level layout plans in the form of scene graphs; (2) a compositional diffusion model that refines the scene graph using estimated object sizes and predicts precise spatial positions; and (3) a foreground-conditioned image generator that synthesizes coherent and realistic scenes by rendering objects according to the planned layout. We formalize the problem addressed by LayoutAgent and provide a detailed breakdown of each component, including implementation specifics and design rationale.

\subsection{Problem Formulation}
The user input to LayoutAgent consists of a natural language description of the target scene (e.g., “in a modern living room”), and $N$ object images $I = \{i_0, i_1, \dots, i_{N-1}\}$ to be placed within the scene. With a Visual-Language Model (VLM) agent, it localize and extract objects $O = \{o_0, o_1, \dots, o_{N-1}\}$ with alpha layer from the images and generates the sizes of objects $S= \{s_0, s_1, \dots, s_{N-1}\}$ including their physical dimensions $(w_i, l_i, h_i)$, which denote width, length, and height (in inches) and other attributes of objects $A= \{a_0, a_1, \dots, a_{N-1}\}$, and constructs a scene graph $G$ that encodes a set of spatial relationships $R$ among objects. This scene graph serves as an intermediate representation for predicting object placements.  Based on the constructed scene graph $G$, VLM agent refine the target scene to detailed natural-language prompts \emph{prompt}. Next, a compositional diffusion model learns the spatial positions of objects $P = \{p_0, p_1, \dots, p_{N-1}\}$, where each $p_i$ corresponds to the bounding box position of object $o_i$.  Finally, a composed image $C$ integrates the given objects and the generated background.  

In LayoutAgent, the task of generating object poses for a given scene is decomposed into four subproblems: (1) processing input object images and representing them as objects; (2) estimating object sizes and constructing a scene graph with spatial relationships among objects and refine prompt based on the generated scene graph; (3) predicting object positions conditioned on the scene graph; and (4) generating a composed image with a background guided by the predicted positions. 

\subsection{Vision-Language Agent for Scene Planning}
Given an abstract scene specification and a set of object images, the goal is to produce coherent final compositions with realistic backgrounds. This task is inherently challenging because scene descriptions are often ambiguous and object images are unprocessed. 
To address the randomness and flexibility of the input, we introduce a \textbf{VLM agent} that automates the scene planning process and enables precise spatial reasoning. 

Concretely, to transform raw object inputs and abstract scene descriptions into coherent compositions, our vision-language agent need to decompose the scene planning into four sub-tasks and complete in a sequence:  1.) \textbf{object extraction}, where the VLM agent utilize tools to localize relevant objects $O$ from images with an open-vocabulary detector (Grounding-DINO~\cite{liu2024grounding}), and applies segmentation (Segment-Anything~\cite{kirillov2023segment} for multi-object images or BiRefNet~\cite{zheng2024bilateral} for single-object images) followed by cropping and overlaying to produce clean objects. 2.) Building on these extracted inputs, VLM agent performs \textbf{attribute estimation}, inferring geometric and appearance attributes $A$ such as real sizes $S$, aspect ratio, and texture and. These attributes serve as symbolic constraints that guide composition. 3.) The agent then performs \textbf{spatial planning}, jointly reasoning over the scene description and the attribute-augmented object set to infer plausible layouts. This process is formalized as a scene graph $G$ that captures both pairwise relations and global structure, serving as an explicit guidance for downstream composition. To ensure fidelity, the scene graph is iteratively evaluated and refined through inspection of the visualization tool. 4.) VLM agent performs \textbf{prompt refinement}, integrating the scene graph with the original description to produce detailed natural-language prompts \emph{prompt}. These refined prompts incorporate object attributes and spatial relations, yielding fine-grained background descriptions for the final image composition.

\subsection{Compositional Diffusion for Layout Generation}
Directly utilizing VLM to generate precise position of objects given complex scene graph and infer their relations
has two main issues: first, for VLM based prediction, due to highly complexity of scene graph and lack of capacity of spatial understanding, VLM tends to predict unprecise location like floating objects and unreal relative size; second, different sizes and scenes typically require a lot of training data for optimal model performance. 

To address the aforementioned issues, we employ compositional diffusion, which leverages individual diffusion-based generative models for each relation type ${r \in R}$ and combines their respective denoising gradients to sample $P$ from high-scoring regions of the distribution~\cite{yang2023compositional}. Given a set of relationships $R$, position of objects $P$, and size of objects $S$, our goal is to find a position $P_0$ that satisfies all relationships. We model the conditional distribution of the position under a single relationship $r$ using a diffusion model. A valid assignment $P_0$ corresponds to the maximizer of the joint distribution formed by the product of all single diffusion models. Each single diffusion model takes the form of an energy-based model,
\[
p(P_r \mid S_r) \propto e^{-E(P_r \mid S_r)},
\]
so maximizing the joint likelihood reduces to minimizing the total energy:
\[
P_0 = \arg\max_P \prod_{r \in R} p_c(P_r \mid S_r) 
\;=\; \arg\min_P \sum_{r \in R} E(P_r \mid S_r).
\]

We solve this optimization using the annealed unadjusted Langevin algorithm (ULA)~\cite{welling2011bayesian, song2020score}, a variant of Langevin Monte Carlo that updates samples by following the gradient of the energy function while injecting noise at each step.

During training, we learn a noise-prediction model $\epsilon_t$ for each relationship $r$. Given sizes of objects $S_r$ and a noisy version of $P_r$, $\epsilon_t$ predicts the applied noise. While one could train individual diffusion models for each relationship, we instead train them jointly: each training example corresponds to a scene graph, and we optimize all instantiated relationships together. Importantly, training and testing graphs do not need to match exactly; they only need to be composed of the same basic relationships.

Formally, our dataset consists of scene graphs with object sizes and corresponding positions of each object, 
$\langle G=(S,P,R), S, P_0 \rangle$. For each relationship, we initialize a denoising function $\epsilon_t$, and compute the denoising loss by summing over relationships in the graph:
\[
\mathcal{L}_{\text{MSE}} = 
\mathbb{E}_{\langle G=(P,S,R), S, P_0 \rangle \in D}
\; \mathbb{E}_{\epsilon \sim \mathcal{N}(0,I), \; t \sim \text{Unif}(1,\ldots,T)}
\left\|
\epsilon - \sum_{r \in R} 
\epsilon_{t_r}\!\left(
\sqrt{\bar{\alpha}_t} P_r^0 + \sqrt{1-\bar{\alpha}_t}\,\epsilon, S_c, t
\right)
\right\|^2 .
\]

\subsection{Final Scene Generation}
To construct the final composed images, we adopt a two-stage generation process. In the first stage, we place each object instance into its predicted bounding box on a uniform background canvas, thereby ensuring that the spatial layout faithfully reflects the structural plan inferred by compositional diffusion model. This intermediate composition serves as a spatially grounded scaffold for subsequent generation. In the second stage, to achieve foreground-conditioned image generation, we leverage a foreground-conditioned image generation model that takes the structured layout as input and renders coherent, high-fidelity images in which objects and background are seamlessly integrated.
\section{Experiments}

In this section, we provide an extensive evaluation of our framework to verify its effectiveness. An ablation study is conducted to demonstrate the effect of compositional diffusion models compared to VLM prediction. We also test generalization of our framework under novel scenes and different objects. Then we showcase qualitative results and application scenarios of our method.

\subsection{Experiment Setup}
\label{sec:metrics}

\textbf{Baselines.} Since we focus on layout generation task, we compare our framework against state-of-the-art frameworks: LayoutGPT ~\cite{feng2023layoutgpt} with different VLMs (GPT-4o and Gemini-2.5), an image layout framework; SpatialRGPT ~\cite{cheng2024spatialrgpt}, a grounded spatial reasoning enhanced VLM; Visual sketchpad ~\cite{hu2024visual}, a visual tools to sketch image thinking in VLM; our layoutAgent with different VLMs(GPT-4o and Gemini-2.5). 

\textbf{Training Setup.} 
We use GPT-4o\footnote{https://openai.com/index/hello-gpt-4o/} and Gemini-2.5\footnote{https://blog.google/technology/google-deepmind/gemini-model-thinking-updates-march-2025/} as the VLM and Imagen\footnote{https://deepmind.google/models/imagen/} for background generation model. Compositional diffusion models are implemented with PyTorch. Experiments are run on two L4 GPUs. We adopt 1000 diffusion timesteps for all models. For input encoding, object dimensions and positions are normalized with respect to the image size. Our compositional diffusion model employs separate encoders for object sizes and positions, projecting them into a shared 256-dimensional latent space, consistent with the time embedding. Each diffusion model produces outputs of the same dimensionality as the position embedding. For each object, the predicted position is obtained by averaging the outputs of all diffusion models in which the object appears, followed by a position decoder that reconstructs the noise added to the position. The encoders, diffusion models, and the position decoder are jointly trained using an L2 reconstruction loss between predicted and ground-truth noise. 

Since our experiments focus on indoor scenes, we collected 300 training images from a design website, covering a variety of interior settings. For each image, we manually label the attributes of each main object, and utilize GroundingDINO~\cite{liu2024grounding} to detect the position of each object to train the compositional diffusion model.

\textbf{Evaluation Setup.}
Our framework is capable of generating both indoor and outdoor scenes. However, since the compositional diffusion models are trained exclusively on indoor scenes, we treat outdoor scenes as out-of-domain data to evaluate the generalization ability of our framework. Specifically, we evaluate our framework across four indoor scene categories: living room, bedroom, billiard room, and a combined warehouse/garage category (due to their similarity in object settings). For each category, we manually select 50 object combinations containing 2 to 4 objects. For each combination, we generate three distinct composite images to compute the evaluation metrics.

\textbf{Metrics.} To evaluate each component of our framework comprehensively, we design metrics scores for each module and overall framework. Specifically, to evaluate LayoutAgent for scene planning, we adopt two metrics: the \textbf{success rate}, which measures whether the VLM agent can generate the required images and scene graphs with estimated attributes for subsequent modules, and the \textbf{IoU score}~\cite{rezatofighi2019generalized}, which assesses the accuracy of the estimated object sizes in Section~\ref{subsec:Agent}. Besides, to evaluate the scene graph built by LayoutAgent, we propose \textbf{relationship coverage} measuring the rate of relations in each scene graph, and \textbf{degree of each nodes} measuring the average edges per node in a scene graph to evaluate the complexity and expressiveness of scene graphs. We also evaluate \textbf{conflict of selected relationships}(`in',`in-front-of',`away-from',`close-to',`left-of') in the scene graph. For example, if ( A, `left-of' B) exist, then ( B, `left-of', A) lead to conflict, and if (A, `close-to', B) exist, (A, `away-from', B) lead to conflict. 

To evaluate the predicted positions (bounding boxes) following scene graphs, we evaluate each relationship in the generated scene graphs following the corresponding rules shown in Table~\ref{tab1:rule}. The final score is calculated by the percentage of rules satisfied.

To quantitatively evaluate the visual quality and spatial realism , we use three metrics: \textbf{CLIP Score}~\cite{radford2021learning} measuring cosine similarity between image and text features from CLIP~\cite{radford2021learning}, \textbf{BLIP Score}~\cite{li2023blip} measuring image-text alignment
using BLIPv2 model and \textbf{VQA Score}~\cite{lin2024evaluating} measuring image caption alignment, which utilizes a VLM to evaluate the final generated image.

Since visual quality and spatial realism are also subjective, we let VLM to judge the final image based on three aspects: (1) \textbf{Layout Coherence} – whether object positions and orientations follow common sense.
(2) \textbf{Spatial Realism} – the extent to which diverse spatial relationships are present (e.g., on top of, inside, under).
(3) \textbf{Aesthetic Alignment} – how well object categories and visual styles align with the scene type. (4) \textbf{overall performance}: the overall visual performance from the generated images.  

\begin{center}
\small
\setlength{\tabcolsep}{6pt}
\captionof{table}{Unary and binary spatial relationships and their implementations.}
\renewcommand{\arraystretch}{1.2}
\begin{tabular}{p{0.18\linewidth} p{0.32\linewidth} p{0.42\linewidth}}
\toprule
\textbf{Relationship} & \textbf{Description} & \textbf{Implementation of $h_R$} \\
\midrule
in(Obj\_A, scene) & Obj\_A must lie completely inside the image boundary. & The four corners of Obj\_A's bounding box must be within the background size (e.g., 1024 $\times$ 1024). \\
\midrule
right-in(Obj\_A, scene) & Obj\_A lies in the right half of scene. & The right edge of Obj\_A is within the right boundary of scene and also within the right half of scene. \\
\midrule
left-in(Obj\_A, Obj\_B) & Obj\_A lies in the left half of scene. & The left edge of Obj\_A is within the left boundary of scene and also within the left half of scene. \\
\midrule
in(Obj\_A, Obj\_B) & Obj\_A is fully contained within Obj\_B. & All edges of Obj\_A are inside the boundaries of Obj\_B. \\
\midrule
left-of(Obj\_A, Obj\_B) & Obj\_A is to the left of Obj\_B. & The horizontal center of Obj\_A is less than the horizontal center of Obj\_B. \\
\midrule
top-of(Obj\_A, Obj\_B) & Obj\_A is above Obj\_B. & The vertical center of Obj\_A is less than the vertical center of Obj\_B. \\
\midrule
close-to(Obj\_A, Obj\_B) & Obj\_A is spatially close to Obj\_B. & The Euclidean distance between the centers of Obj\_A and Obj\_B is smaller than a threshold. \\
\midrule
away-from(Obj\_A, Obj\_B) & Obj\_A is spatially far from Obj\_B. & The Euclidean distance between the centers of Obj\_A and Obj\_B is larger than a threshold. \\
\midrule
overlapping(Obj\_A, Obj\_B) & Obj\_A and Obj\_B overlap in space. & The intersection of their bounding boxes has positive width and height. \\
\midrule
in-front-of(Obj\_A, Obj\_B) & Obj\_A appears in front of Obj\_B (vertical ordering). & The vertical center of Obj\_A is smaller (higher up) than the vertical center of Obj\_B. \\
\bottomrule
\end{tabular}
\label{tab1:rule}
\end{center}

\subsection{Main Results}

Tables~\ref{tab:all} compares scene graph generation, positional accuracy, visual-text alignment and VLM based evaluation across baseline methods.
Plain VLM-based approaches, which are prompted to directly produce scene graphs and bounding boxes, achieve only modest scores across all dimensions. Their outputs often exhibit incomplete relational structures and imprecise spatial grounding, which in turn leads to weaker performance in downstream evaluations of layout coherence and realism. This suggests that direct prompting is insufficient for capturing the complexity of multi-object interactions.

LayoutGPT provides an improvement by retrieving similar layout as few-show learning. While this yields better relational coverage than plain VLM prompting, the resulting relationships between objects remain shallow and the associated bounding boxes exhibit frequent mis-alignments in complex scenes. Consequently, while LayoutGPT produces moderate gains in text-image alignment, its overall image quality metrics remain limited by the lack of robust spatial reasoning.

SpatialRGPT and Visual Sketchpad both enhance text–image alignment as well as layout coherence and spatial realism under VLM-based evaluation. SpatialRGPT, by emphasizing spatial reasoning, yields higher spatial realism than plain VLMs; however, it remains inferior to our LayoutAgent across these metrics, suggesting that spatial reasoning alone is insufficient for producing satisfactory layouts. Visual Sketchpad benefits from its use of visual tools, which improves overall scores, yet in the absence of scene graphs and compositional diffusion models it struggles to reliably capture fine-grained spatial positions and generate coherent layouts. 

Our LayoutAgent leverages the spatial realism by explicitly conditioning on structured scene graphs and relational constraints. It substantially improves bounding box accuracy, producing layouts that are both coherent and spatially consistent across diverse scenes, which proves the effectiveness of compositional diffusion models on precise layout generation. VLM-based evaluation further confirms the advantage: compared to baselines, scene graphs generated by our visual-language agent achieves significantly higher scores on layout coherence, spatial realism, and overall quality. Importantly, the improvements in aesthetic alignment stem from our agent modules, including prompt refinement and background generation, which ensure that scene structure and visual fidelity are jointly optimized.

Collectively, these results demonstrate that integrating a VLM agent generating structured scene graph representations with diffusion-based composition yields the most reliable and diverse scene layouts, outperforming prompt-driven and tool-only baselines on both structural and perceptual dimensions.

\begin{table}[t]
\centering
\small
\setlength{\tabcolsep}{4pt} 
\renewcommand{\arraystretch}{1.1} 
\caption{Quantitative evaluation on baselines and our LayoutAgent
($\uparrow$ higher is better). Bold marks the best for text control measurement.}
\label{tab:all}
\resizebox{\linewidth}{!}{ 
\begin{tabular}{lccc|c|ccc|cccc}
\toprule
\multirow{2}{*}{Method} 
  & \multicolumn{3}{c|}{Scene Graph} 
  & \multirow{2}{*}{Pos. Score} 
  & \multicolumn{3}{c|}{Text--Image Align} 
  & \multicolumn{4}{c}{VLM Eval} \\
\cmidrule(lr){2-4} \cmidrule(lr){6-8} \cmidrule(lr){9-12}
 & Rel Cov $\uparrow$ & Deg $\uparrow$ & Conf $\downarrow$
 &  & CLIP $\uparrow$ & BLIP $\uparrow$ & VQA $\uparrow$
 & Layout Coh $\uparrow$ & Spatial $\uparrow$ & Aest Align $\uparrow$ & Overall $\uparrow$ \\
\midrule
Plain GPT-4o        & - & - & - & 0.37 & 20.12 & 54.23 & 0.43 & 0.47 & 0.44 & 0.46 & 0.45 \\ 
Plain Gemini-2.5     & - & - & - & 0.36 & 19.69 & 53.43 & 0.41 & 0.45 & 0.42 & 0.47 & 0.44 \\ 
Plain Gemini-2.5pro  & - & - & - & 0.39 & 21.05 & 54.78 & 0.42 & 0.48 & 0.43 & 0.49 & 0.51 \\ 
LayoutGPT + GPT-4o        & - & - & - & 0.51 & 23.77 & 62.10 & 0.61 & 0.62 & 0.61 & 0.63 & 0.60 \\ 
LayoutGPT + Gemini-2.5   & - & - & - & 0.49 & 23.20 & 61.72 & 0.59 & 0.58 & 0.61 & 0.61 & 0.57 \\
LayoutGPT + Gemini-2.5pro & - & - & - & 0.52 & 24.32 & 62.48 & 0.61 & 0.60 & 0.62 & 0.64 & 0.61 \\
SpatialRGPT            & - & - & - & 0.54 & 26.52 & 64.32 & 0.67 & 0.63 & 0.68 & 0.69 & 0.62 \\
Visual Sketchpad           & - & - & - & 0.52 & 25.80 & 63.50 & 0.66 & 0.64 & 0.66 & 0.68 & 0.63 \\
LayoutAgent + GPT-4o             & 0.59 & 2.84 & 0 & 0.68 & 29.74 & 70.40 & 0.79 & 0.71 & 0.75 & 0.76 & 0.71 \\
LayoutAgent + Gemini-2.5         & 0.57 & 2.55 & 0.07 & 0.64 & 28.46  & 68.57  & 0.77 & 0.68 & 0.77 & 0.72 & 0.69 \\
\textbf{LayoutAgent + Gemini-2.5pro} 
                         & \textbf{0.61} & \textbf{3.05} & \textbf{0} 
                         & \textbf{0.72} &  \textbf{31.62} & \textbf{74.23} &\textbf{0.81} 
                         & \textbf{0.72} & \textbf{0.82} & \textbf{0.79} & \textbf{0.73} \\
\bottomrule
\end{tabular}}
\end{table}

\subsection{Ablation Study}
In our framework, two key components drive performance: scene graph construction and the compositional diffusion model. To assess their contributions, we conduct ablation studies. Since these experiments isolate the effect of each module on the final image quality, we adopt the text-alignment metrics and VLM-based evaluation introduced in Section~\ref{sec:metrics}. As shown in Table~\ref{tab:abla}, the scene graph significantly improves text alignment, layout coherence, and aesthetics alignment in the VLM evaluation, demonstrating its ability to encode concrete spatial relationships that align with human preferences. Moreover, the compositional diffusion model further improves spatial realism, highlighting its capacity to generate more precise spatial placements compared to VLM-only methods.

\begin{table}[htbp]
\centering
\scriptsize
\setlength{\tabcolsep}{7pt} 
\renewcommand{\arraystretch}{1.15} 
\caption{Ablation study on VLM evaluation metrics, Scene graphs and Text Alignment Scores.
($\uparrow$ higher is better, $\downarrow$ lower is better). CD is abbreviation of Compositional Diffusion.}
\label{tab:abla}
\begin{tabular}{l|c|rrr|rrr}
\toprule
\multirow{2}{*}{Method} & \multicolumn{1}{c|}{VLM Overall Score} 
& \multicolumn{3}{c|}{Scene Graph} 
& \multicolumn{3}{c}{Text Align Score} \\
\cmidrule(lr){2-2} \cmidrule(lr){3-5} \cmidrule(lr){6-8}
& & Rel Cov $\uparrow$ & Deg $\uparrow$ & Conf $\downarrow$ 
& CLIP $\uparrow$ & BLIP $\uparrow$ & VQA $\uparrow$ \\
\midrule
VLM & 0.56 & - & - & - & 26.54 & 72.45.3 & 0.68 \\
VLM + Scene Graph & 0.62 & 0.61 & 3.05 & 0 & 27.68 & 73.25 & 0.72 \\
VLM + Scene Graph + CD & \textbf{0.73} & \textbf{0.61} & \textbf{3.05} & \textbf{0} & \textbf{31.62} & \textbf{74.23} & \textbf{0.81} \\
\bottomrule
\end{tabular}
\end{table}

\subsection{Generlizability}
In this section, we further analyze generalization across different scene settings, including generalization to novel scenes, as well as more objects to plan.

\textbf{Novel scene}
Since the training data is limited to indoor scenes, we evaluate our framework under an out-of-domain setting by testing on an outdoor beach scene. As shown in Table~\ref{tab:beach}, our framework achieves the best performance among all baselines, with only a minimal decrease compared to in-domain results. This demonstrates that our approach is more robust than the baselines when faced with out-of-domain inputs, suggesting that the scene graph component helps bridge semantic gaps between indoor and outdoor scenes. Furthermore, because the compositional diffusion model does not rely on domain-specific visual cues during training, it exhibits strong generalization across different object sizes and scene types. 

\textbf{Complex Scene with Lots of Objects}
To evaluate our framework in complex scenarios with more objects, we conduct experiments on 4-object scenes and compare the results with baseline methods. As shown in Table~\ref{tab:four-objects}, our framework exhibits the smallest performance degradation as the number of objects increases. This demonstrates the effectiveness of scene graphs in handling richer object interactions and highlights the robustness of compositional diffusion models when generating images with more objects and complex spatial relationships.

\begin{center}
\small
\captionof{table}{Performance comparison for \textbf{Beach Scene}}
\label{tab:beach}
\begin{tabular}{lccc}
\toprule
\textbf{Model} & \textbf{Deg} & \textbf{Pos Score} & \textbf{VLM Overall Score} \\
\midrule
Plain GPT-4o & - & 0.37 & 0.41 \\
Plain Gemini-2.5 & - & 0.36 & 0.40 \\
Plain Gemini-2.5pro & - & 0.36 & 0.42 \\
LayoutGPT + GPT-4o & - & 0.46 & 0.49 \\
LayoutGPT + Gemini-2.5 & - & 0.43 & 0.49 \\
LayoutGPT + Gemini-2.5pro & - & 0.45 & 0.51 \\
SpatialRGPT & - & 0.49 & 0.54 \\
Visual Sketchpad & - & 0.52 & 0.58 \\
LayoutAgent + GPT-4o & 3.0 & 0.63 & 0.67 \\
LayoutAgent + Gemini-2.5 & 2.9 & 0.62 & 0.65 \\
LayoutAgent + Gemini-2.5pro & \textbf{3.1} & \textbf{0.63} & \textbf{0.67} \\
\bottomrule
\end{tabular}
\end{center}

\begin{center}
\small
\captionof{table}{Performance comparison for \textbf{4 Objects Scene}}
\label{tab:four-objects}
\begin{tabular}{lccc}
\toprule
\textbf{Model} & \textbf{Deg} & \textbf{Pos Score} & \textbf{VLM Overall Score} \\
\midrule
Plain GPT-4o & - & 0.35 & 0.41 \\
Plain Gemini-2.5 & - & 0.34 & 0.40 \\
Plain Gemini-2.5pro & - & 0.37 & 0.43 \\
LayoutGPT + GPT-4o & - & 0.48 & 0.52 \\
LayoutGPT + Gemini-2.5 & - & 0.46 & 0.51 \\
LayoutGPT + Gemini-2.5pro & - & 0.47 & 0.51 \\
SpatialRGPT & - & 0.54 & 0.57 \\
Visual Sketchpad & - & 0.51 & 0.55 \\
LayoutAgent + GPT-4o & 3.4 & 0.63 & 0.66 \\
LayoutAgent + Gemini-2.5 & 3.4 & 0.62 & 0.65 \\
LayoutAgent + Gemini-2.5pro & \textbf{3.6} & \textbf{0.64} & \textbf{0.68} \\
\bottomrule
\end{tabular}
\end{center}

\subsection{Agent Completion Analysis}
\label{subsec:Agent}
In this section, we explore the effect of different VLM backbones on our agent framework. We compare GPT-4o, Gemini-2.5, and Gemini-2.5 pro by evaluating complexity and conflict of generated scene graphs, the success rate of completing tasks generating intermediate results for later modules, and IoU ~\cite{rezatofighi2019generalized} between predicted size and real size of objects in weight, height and length. From Table~\ref{tab:agent}, we find Gemini-2.5pro achieves the highest success rate of completing tasks, best performance on scene graph generation and attributes estimation. This implies that Gemini-2.5pro is more suitable for our agent framework compared to other VLMs, and shows better generalized spatial understanding and planning capability. 

\begin{table}[htbp]
\centering
\small
\setlength{\tabcolsep}{4pt}
\renewcommand{\arraystretch}{1.2}
\caption{Comparison of different VLM backbones on scene graph metrics, task success rate, and IoU.}
\label{tab:agent}
\begin{tabular}{lccc|cc}
\toprule
\multirow{2}{*}{Method} & \multicolumn{3}{c|}{Scene Graph} & \multirow{2}{*}{Success Rate $\uparrow$} & \multirow{2}{*}{IoU $\uparrow$} \\
\cmidrule(lr){2-4}
& Rel Cov $\uparrow$ & Deg $\uparrow$ & Conf $\downarrow$ & & \\
\midrule
LayoutAgent + GPT-4o & 0.59 & 2.84 & 0 & 0.91 & 0.75 \\
LayoutAgent + Gemini-2.5 & 0.57 & 2.55 & 0.07 & 0.88 & 72 \\
\textbf{Ours + Gemini-2.5pro)} & \textbf{0.61} & \textbf{3.05} & \textbf{0} & \textbf{0.92} & \textbf{0.78} \\
\bottomrule
\end{tabular}
\end{table}

\subsection{Case Studies}
\begin{figure*}[htbp]
    \centering
    \includegraphics[width=0.9\textwidth]{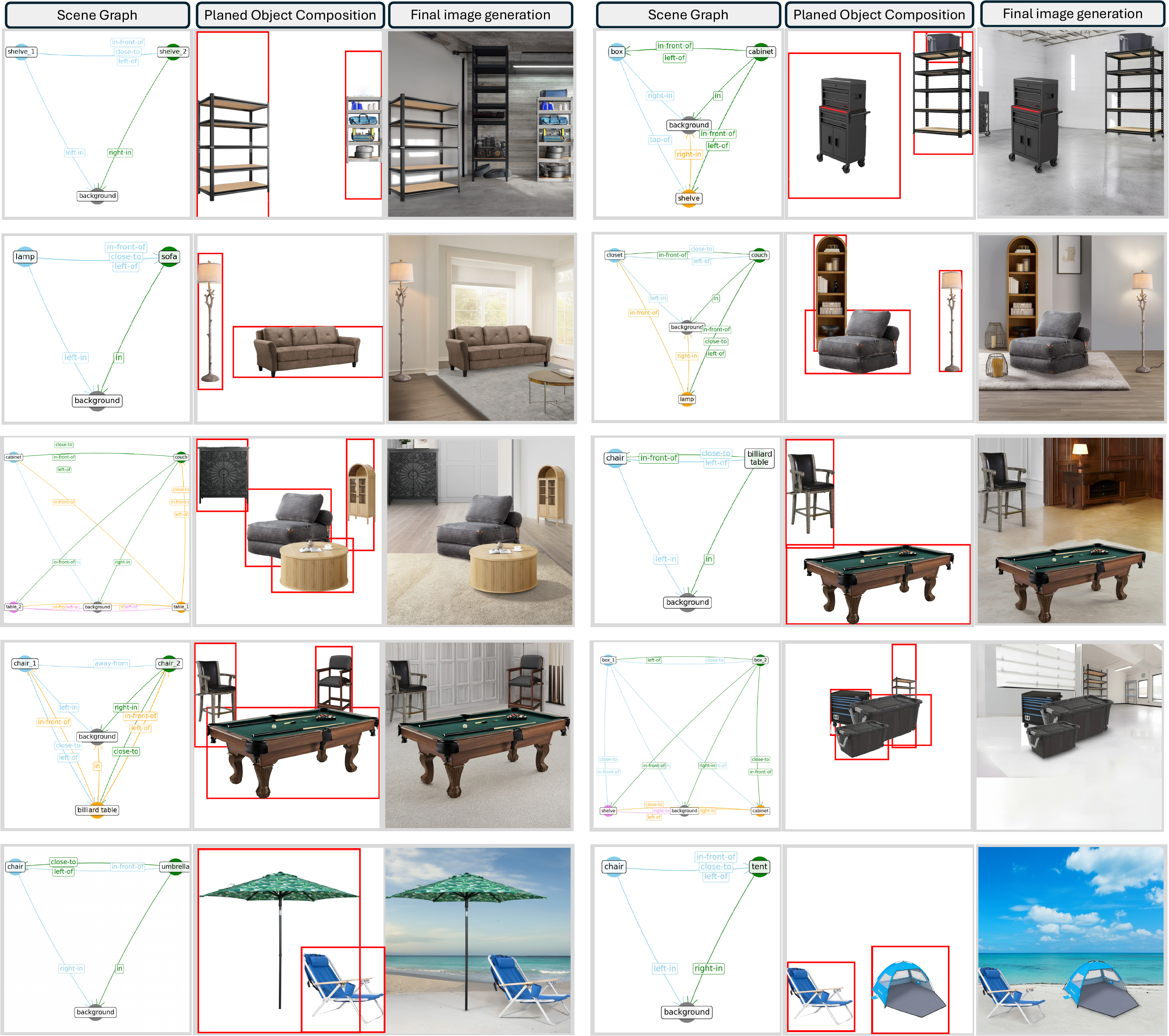}
    \caption{Case studies on the spatial planning and scene generation}
    \label{fig:case_study}
\end{figure*}

To demonstrate the effectiveness of our LayoutAgent, the case studies in Figure~\ref{fig:good_cases} and \ref{fig:case_study} illustrate its systematic, multi-stage planning for scene generation. The process begins when the VLM Agent analyzes the input objects and formulates a coherent plan as a symbolic \textbf{Scene Graph}, which establishes logical spatial relationships between objects. Following this plan, our compositional diffusion model translates these abstract relationships into a precise spatial arrangement, predicting the exact bounding box locations for each object as shown in the \textbf{Planed Object Composition}. Finally, a foreground-conditioned image generator uses this structured layout and a refined prompt from the agent to render the \textbf{Final image generation}, seamlessly integrating the objects by synthesizing a realistic and semantically consistent.
The novel beach scene is outside the training data distribution, which is primarily for indoor settings. Our LayoutAgent can understand the logical spatial relationships between objects and plan the scene layout. This step-by-step approach, validated across diverse indoor and outdoor scenes, shows the unique strength of LayoutAgent in unifying high-level reasoning with precise generative control to produce realistic and well-structured final images.

\section{Conclusion}


We presented LayoutAgent, a vision-language guided compositional diffusion framework that unifies symbolic spatial reasoning with generative modeling for scene planning. Our approach produces spatially plausible and semantically coherent layouts, surpassing conventional planning methods. 

\clearpage
\nocite{*}
\bibliographystyle{plain}
\bibliography{ref}

\begin{thebibliography}{10}

\bibitem{betker2023improving}
James Betker, Gabriel Goh, Li~Jing, Tim Brooks, Jianfeng Wang, Linjie Li, Long
  Ouyang, Juntang Zhuang, Joyce Lee, Yufei Guo, et~al.
\newblock Improving image generation with better captions.
\newblock {\em Computer Science. https://cdn. openai. com/papers/dall-e-3.
  pdf}, 2(3):8, 2023.

\bibitem{brohan2023rt2}
Anthony Brohan et~al.
\newblock Rt-2: Vision-language-action models transfer web knowledge to robotic
  control.
\newblock {\em arXiv preprint arXiv:2307.15818}, 2023.

\bibitem{chen2023program}
Mark Chen et~al.
\newblock Program synthesis with large language models.
\newblock {\em arXiv preprint arXiv:2303.12345}, 2023.

\bibitem{cheng2024spatialrgpt}
An-Chieh Cheng, Hongxu Yin, Yang Fu, Qiushan Guo, Ruihan Yang, Jan Kautz,
  Xiaolong Wang, and Sifei Liu.
\newblock Spatialrgpt: Grounded spatial reasoning in vision-language models.
\newblock {\em Advances in Neural Information Processing Systems},
  37:135062--135093, 2024.

\bibitem{fan2024prompt}
Zezhong Fan, Xiaohan Li, Kaushiki Nag, Chenhao Fang, Topojoy Biswas, Jianpeng
  Xu, and Kannan Achan.
\newblock Prompt optimizer of text-to-image diffusion models for abstract
  concept understanding.
\newblock In {\em Companion Proceedings of the ACM Web Conference 2024}, pages
  1530--1537, 2024.

\bibitem{feng2023layoutgpt}
Weixi Feng, Wanrong Zhu, Tsu-jui Fu, Varun Jampani, Arjun Akula, Xuehai He,
  Sugato Basu, Xin~Eric Wang, and William~Yang Wang.
\newblock Layoutgpt: Compositional visual planning and generation with large
  language models.
\newblock {\em Advances in Neural Information Processing Systems},
  36:18225--18250, 2023.

\bibitem{hu2024visual}
Yushi Hu, Weijia Shi, Xingyu Fu, Dan Roth, Mari Ostendorf, Luke Zettlemoyer,
  Noah~A Smith, and Ranjay Krishna.
\newblock Visual sketchpad: Sketching as a visual chain of thought for
  multimodal language models.
\newblock {\em Advances in Neural Information Processing Systems},
  37:139348--139379, 2024.

\bibitem{jiang2023mistral}
Zhengbao Jiang et~al.
\newblock Mistral: Efficient transformer model for language understanding.
\newblock {\em arXiv preprint arXiv:2309.12345}, 2023.

\bibitem{kirillov2023segment}
Alexander Kirillov, Eric Mintun, Nikhila Ravi, Hanzi Mao, Chloe Rolland, Laura
  Gustafson, Tete Xiao, Spencer Whitehead, Alexander~C Berg, Wan-Yen Lo, et~al.
\newblock Segment anything.
\newblock In {\em Proceedings of the IEEE/CVF international conference on
  computer vision}, pages 4015--4026, 2023.

\bibitem{li2023blip}
Junnan Li, Dongxu Li, Silvio Savarese, and Steven Hoi.
\newblock Blip-2: Bootstrapping language-image pre-training with frozen image
  encoders and large language models.
\newblock In {\em International conference on machine learning}, pages
  19730--19742. PMLR, 2023.

\bibitem{lin2024evaluating}
Zhiqiu Lin, Deepak Pathak, Baiqi Li, Jiayao Li, Xide Xia, Graham Neubig,
  Pengchuan Zhang, and Deva Ramanan.
\newblock Evaluating text-to-visual generation with image-to-text generation.
\newblock In {\em European Conference on Computer Vision}, pages 366--384.
  Springer, 2024.

\bibitem{liu2022compositional}
Nan Liu, Shuang Li, Yilun Du, Antonio Torralba, and Joshua~B Tenenbaum.
\newblock Compositional visual generation with composable diffusion models.
\newblock In {\em European conference on computer vision}, pages 423--439.
  Springer, 2022.

\bibitem{liu2024grounding}
Shilong Liu, Zhaoyang Zeng, Tianhe Ren, Feng Li, Hao Zhang, Jie Yang, Qing
  Jiang, Chunyuan Li, Jianwei Yang, Hang Su, et~al.
\newblock Grounding dino: Marrying dino with grounded pre-training for open-set
  object detection.
\newblock In {\em European conference on computer vision}, pages 38--55.
  Springer, 2024.

\bibitem{mialon2023augmented}
Grégoire Mialon et~al.
\newblock Augmented language models: a survey.
\newblock {\em arXiv preprint arXiv:2302.07842}, 2023.

\bibitem{openai2023gpt4}
OpenAI.
\newblock Gpt-4 technical report.
\newblock {\em arXiv preprint arXiv:2303.08774}, 2023.

\bibitem{openai2023gpt4v}
OpenAI.
\newblock Gpt-4v: Vision-language capabilities of gpt-4.
\newblock {\em arXiv preprint arXiv:2303.08774}, 2023.

\bibitem{radford2021learning}
Alec Radford, Jong~Wook Kim, Chris Hallacy, Aditya Ramesh, Gabriel Goh,
  Sandhini Agarwal, Girish Sastry, Amanda Askell, Pamela Mishkin, Jack Clark,
  et~al.
\newblock Learning transferable visual models from natural language
  supervision.
\newblock In {\em International conference on machine learning}, pages
  8748--8763. PmLR, 2021.

\bibitem{ramesh2023dalle}
Aditya Ramesh et~al.
\newblock Dall·e 3: Language-based image generation with human-like quality.
\newblock {\em arXiv preprint arXiv:2309.12345}, 2023.

\bibitem{rezatofighi2019generalized}
Hamid Rezatofighi, Nathan Tsoi, JunYoung Gwak, Amir Sadeghian, Ian Reid, and
  Silvio Savarese.
\newblock Generalized intersection over union: A metric and a loss for bounding
  box regression.
\newblock In {\em Proceedings of the IEEE/CVF conference on computer vision and
  pattern recognition}, pages 658--666, 2019.

\bibitem{ritchie2019fast}
Daniel Ritchie, Kai Wang, and Yu-an Lin.
\newblock Fast and flexible indoor scene synthesis via deep convolutional
  generative models.
\newblock In {\em Proceedings of the IEEE/CVF conference on computer vision and
  pattern recognition}, pages 6182--6190, 2019.

\bibitem{roberts1996exponential}
Gareth~O Roberts and Richard~L Tweedie.
\newblock Exponential convergence of langevin distributions and their discrete
  approximations.
\newblock 1996.

\bibitem{rombach2022high}
Robin Rombach, Andreas Blattmann, Dominik Lorenz, Patrick Esser, and Bj{\"o}rn
  Ommer.
\newblock High-resolution image synthesis with latent diffusion models.
\newblock In {\em Proceedings of the IEEE/CVF conference on computer vision and
  pattern recognition}, pages 10684--10695, 2022.

\bibitem{song2020score}
Yang Song, Jascha Sohl-Dickstein, Diederik~P Kingma, Abhishek Kumar, Stefano
  Ermon, and Ben Poole.
\newblock Score-based generative modeling through stochastic differential
  equations.
\newblock {\em arXiv preprint arXiv:2011.13456}, 2020.

\bibitem{touvron2023llama}
Hugo Touvron et~al.
\newblock Llama 2: Open foundation and fine-tuned chat models.
\newblock {\em arXiv preprint arXiv:2307.09288}, 2023.

\bibitem{touvron2023llama2}
Hugo Touvron, Louis Martin, Kevin Stone, et~al.
\newblock Llama 2: Open foundation and fine-tuned chat models.
\newblock {\em arXiv preprint arXiv:2307.09288}, 2023.

\bibitem{wang2019planit}
Kai Wang, Yu-An Lin, Ben Weissmann, Manolis Savva, Angel~X Chang, and Daniel
  Ritchie.
\newblock Planit: Planning and instantiating indoor scenes with relation graph
  and spatial prior networks.
\newblock {\em ACM Transactions on Graphics (TOG)}, 38(4):1--15, 2019.

\bibitem{wei2022chain}
Jason Wei et~al.
\newblock Chain-of-thought prompting elicits reasoning in large language
  models.
\newblock {\em NeurIPS}, 2022.

\bibitem{welling2011bayesian}
Max Welling and Yee~W Teh.
\newblock Bayesian learning via stochastic gradient langevin dynamics.
\newblock In {\em Proceedings of the 28th international conference on machine
  learning (ICML-11)}, pages 681--688, 2011.

\bibitem{xu2025set}
Yiqing Xu, Jiayuan Mao, Linfeng Li, Yilun Du, Tomas Loz{\'a}no-P{\'e}rez,
  Leslie~Pack Kaelbling, and David Hsu.
\newblock " set it up": Functional object arrangement with compositional
  generative models.
\newblock {\em arXiv preprint arXiv:2508.02068}, 2025.

\bibitem{yang2023compositional}
Zhutian Yang, Jiayuan Mao, Yilun Du, Jiajun Wu, Joshua~B Tenenbaum, Tom{\'a}s
  Lozano-P{\'e}rez, and Leslie~Pack Kaelbling.
\newblock Compositional diffusion-based continuous constraint solvers.
\newblock In {\em Conference on Robot Learning}, pages 3242--3265. PMLR, 2023.

\bibitem{yao2023tree}
Shunyu Yao et~al.
\newblock Tree of thoughts: Deliberate problem solving with large language
  models.
\newblock {\em arXiv preprint arXiv:2305.10601}, 2023.

\bibitem{yu2022scaling}
Jiahui Yu, Yuanzhong Xu, Jing~Yu Koh, Thang Luong, Gunjan Baid, Zirui Wang,
  Vijay Vasudevan, Alexander Ku, Yinfei Yang, Burcu~Karagol Ayan, et~al.
\newblock Scaling autoregressive models for content-rich text-to-image
  generation.
\newblock {\em arXiv preprint arXiv:2206.10789}, 2(3):5, 2022.

\bibitem{zheng2024bilateral}
Peng Zheng, Dehong Gao, Deng-Ping Fan, Li~Liu, Jorma Laaksonen, Wanli Ouyang,
  and Nicu Sebe.
\newblock Bilateral reference for high-resolution dichotomous image
  segmentation.
\newblock {\em arXiv preprint arXiv:2401.03407}, 2024.

\end{thebibliography}

\end{document}